\documentclass{article} 
\usepackage{nips14submit_e,times} 
\usepackage{graphicx}
\usepackage{fancyvrb}
\usepackage{color}
\usepackage{amsmath}
\usepackage{subcaption}
\usepackage{authblk,subfig,caption} 

\newcommand\blfootnote[1]{%
  \begingroup
  \renewcommand\thefootnote{}\footnote{#1}%
  \addtocounter{footnote}{-1}%
  \endgroup
}

\title{Inverse Graphics with Probabilistic CAD Models}
\author[1,2]{Tejas D. Kulkarni}%
\author[1,2]{Vikash K. Mansinghka}%
\author[3]{Pushmeet Kohli}%
\author[1,2]{Joshua B. Tenenbaum}
\affil[1]{Computer Science and Artificial Intelligence Laboratory, MIT}
\affil[2]{Department of Brain and Cognitive Sciences, MIT}
\affil[3]{Microsoft Research Cambridge
}

\nipsfinalcopy 

\begin{document}
\maketitle
\begin{abstract}
Recently, multiple formulations of vision problems as probabilistic inversions of generative models based on computer graphics have been proposed. However, applications to 3D perception from natural images have focused on low-dimensional latent scenes, due to challenges in both modeling and inference. Accounting for the enormous variability in 3D object shape and 2D appearance via realistic generative models seems intractable, as does inverting even simple versions of the many-to-many computations that link 3D scenes to 2D images. This paper proposes and evaluates an approach that addresses key aspects of both these challenges. We show that it is possible to solve challenging, real-world 3D vision problems by approximate inference in generative models for images based on rendering the outputs of probabilistic CAD (PCAD) programs. Our PCAD object geometry priors generate deformable 3D meshes corresponding to plausible objects and apply affine transformations to place them in a scene. Image likelihoods are based on similarity in a feature space based on standard mid-level image representations from the vision literature. Our inference algorithm integrates single-site and locally blocked Metropolis-Hastings proposals, Hamiltonian Monte Carlo and discriminative data-driven proposals learned from training data generated from our models. We apply this approach to 3D human pose estimation and object shape reconstruction from single images, achieving quantitative and qualitative performance improvements over state-of-the-art baselines.
\end{abstract}

\section{Introduction}

\renewcommand{\thefootnote}{\roman{footnote}}

Formulations of vision problems as probabilistic inversions of generative models based on computer graphics have a long history \cite{baumgart1974geometric,yuille2006vision} and have recently attracted renewed attention \cite{mansinghka2013approximate, jampani2014informed, gupta2010blocks}. However, applications to 3D object perception from single natural images have seemed intractable. Generative models for natural images have instead either focused on 2D problems \cite{ZhuowenZ02}, considered only low-dimensional latent scenes comprised of simple shapes\cite{xiao2012localizing, mansinghka2013approximate}, or made heavy use of temporal continuity \cite{zuffi2013estimating}.
\blfootnote{Contact(left to right): \{tejask@mit.edu\}, \{vkm@mit.edu\}, \{pkohli@microsoft.com\}, \{jbt@mit.edu\}}

On the modeling side, accounting for the enormous variability in 3D object shape and 2D appearance via realistic generative models can seem wildly intractable. The failure of the Inverse Graphics approach has primarily been due to the lack of a {\em generic} graphics engine and the computational intractability of the inversion problem. It has also proved difficult to navigate the tension between model flexibility and inference tractability. Consider that permissive models such as generic priors on 3D meshes both exacerbate the intractability of inference and sometimes lead to unrealistic percepts. On the other hand, identifying high likelihood inputs to a rendering engine can seem challenging enough without additionally accounting for a highly structured scene prior.


This paper proposes and evaluates a new approach that aims to address these challenges. We show that it is possible to solve challenging, real-world 3D vision problems by approximate inference in generative image models for deformable 3D meshes. Our approach uses scene representations that build on tools from computer-aided design (CAD) and nonparametric Bayesian statistics~\cite{rasmussen2006gaussian}. We specify priors on object geometry using {\em probabilistic CAD (PCAD) programs} within a generic rendering engine environment~\cite{blender}: stochastic procedures that sample meshes from component priors and apply affine transformations to place them in a scene. Image likelihoods are based on similarity in a feature space based on standard mid-level image representations from the vision literature~\cite{dollar2013structured, martin2004learning}.

To the best of our knowledge, our system is the first real-world formulation to define rich generative models over rigid and deformable 3D CAD programs to interpret real images. We apply this approach to 3D human pose estimation and object shape reconstruction from single images, achieving quantitative performance that exceeds state-of-the-art baselines. The 3D mesh based parametrization of our model consists of a large set of mixed discrete and continuous latent variables. This, coupled with the complex many-to-many nature of graphics mapping, makes computation of the inverse mapping an extremely difficult inference problem. Our inference algorithm integrates single-site and locally blocked Metropolis-Hastings proposals and Hamiltonian Monte Carlo, and discriminative proposals learned from training data generated from our models. We show that discriminative proposals aids the inference pipeline both in terms of speed and accuracy, allowing us to successfully leverage strengths of discriminative methods in generative modeling.



\begin{figure*}
\centering
\includegraphics[width=70mm]{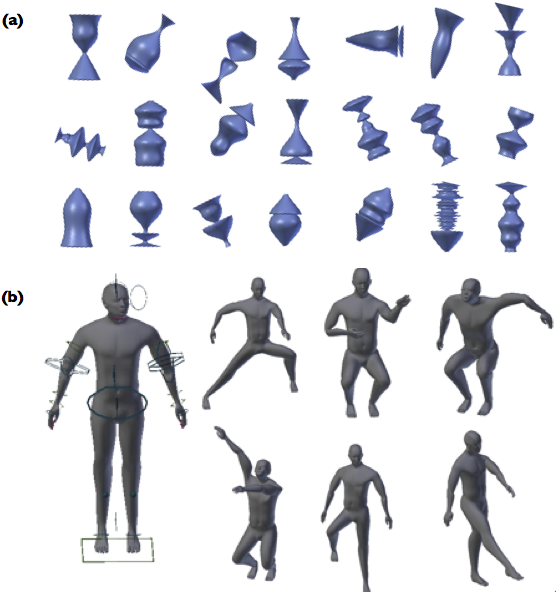}
\includegraphics[width=70mm]{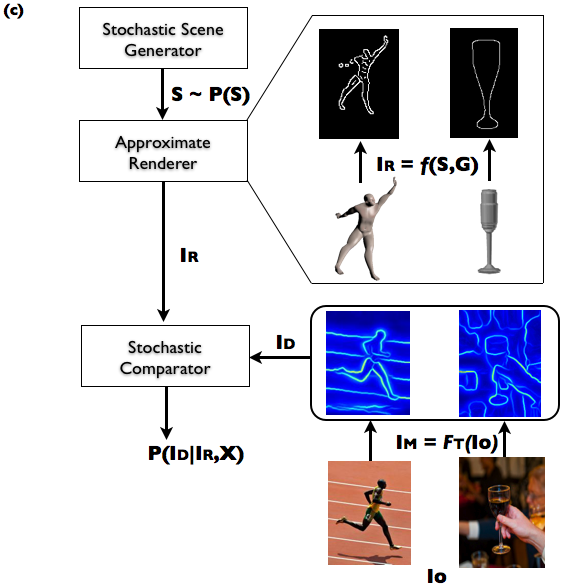}
\caption{{\bf Probabilistic CAD Framework:} Stochastic Scene Generator ({\it SSG}) defines the distribution over 3D mesh and affine latent variables. We use a generic rendering engine~\cite{blender} as the common {\it SSG} to express all our Probabilistic CAD programs by just changing the scene priors $S$. (a) Random samples drawn from our generic 3D Object CAD Model, where the scene prior consists of Gaussian Process and affine transformations. (b) Random samples drawn from the 3D Human CAD model, where the scene priors are defined over the armature of the 3D mesh for smooth deformations and over the affine transformations. (c) Probabilistic CAD models and the scene priors can be viewed as the Stochastic Scene Generator ({\it SSG}). Samples generated from the {\it SSG} are rendered in the Approximate Renderer in form of a mid-level representation and compared with the data via the likelihood function $P(S|I_D)$}
\label{fig:overview}
\end{figure*}

\begin{figure*}
\centering
\includegraphics[height=70mm]{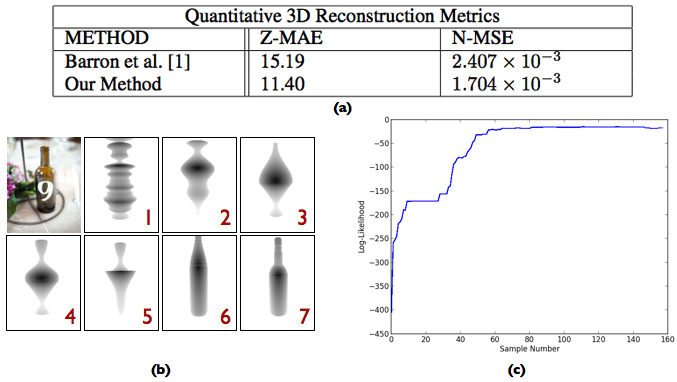}
\caption{{\bf Quantitative Results for 3D Object Parsing:} (a) Quantitative comparison for 3D reconstruction of our method with SIRFS~\cite{barron2013shape}. Our method beats the baseline by a considerable margin without using pre-segmented objects. Moreover, a lot of our quantitative errors can be attributed to minor misalignment in model fit due to lack of segmented object as is given to the baseline. (b,c) Illustration of inference trajectory for a typical run escaping many local minimas. The images depict the z-buffer of the rendered mesh. Image 1 is closer to the prior than Image 7. The prior starts with latents on the GP favoring complicated objects but quickly converges to the posterior as inference progresses.}
\label{fig:gp_quantitative}
\end{figure*}

\section{Modeling via Probabilistic CAD Programs}
Probabilistic deformable CAD programs (PCADs) define generative models for 3D shapes by combining three major components as first proposed in~\cite{mansinghka2013approximate}:

{\noindent {\bf (a) Stochastic scene generator}} is the distribution over 3D meshes and other scene elements such as affine transformations. The scene generator can be factorized into several scene elements $\{ S_i \in S, 1 \leq i \leq N\}$, which denotes scene configuration or latent variables over 3D meshes. This factorization induces a set of latent variables $S_i$ with priors $P(S) = \prod_i P(S_i)$.

{\noindent {\bf (b) Approximate renderer }} projects the 3D mesh to 2D mid-level representation for stochastic comparison. The approximate renderer is a complex simulator denoted by function $I_R = g(\Theta*S)$, where $I_R$ is the 2D projection of the 3D generated scene and $\Theta$ denotes additional control variables for the rendering engine. One of our key contributions is to formulate image interpretation as inverting this simulator under observations. Blender is a widely used open source 3D graphics environment and it exposes numerous interfaces to model, render and manipulate complex 3D scenes. In this paper, we take the idea of inverse graphics literally by abstracting the simulator to be our function $g(.)$ and driving it by putting rich priors over scene elements $S$. In order to abstract away the extreme pixel variability in real world images, we transform RGB images to contour maps with the recent Structured Random Forest model~\cite{dollar2013structured}. The approximate renderer outputs the same mid-level representation conditioned on $S$, resulting in an image hypothesis space with simple appearances variability while preserving 3D geometric details.

{\noindent {\bf (c) Stochastic Comparator}} is a likelihood function $P(I_D|I_R)$ or in case of likelihood free inference~\cite{csillery2010approximate}, a distance function $\rho(I_D,I_R)$. In our experiments, we use a contour based representation for both data $I_D$ and the rendered image $I_R$. Our stochastic comparator is defined as a probabilistic variant of the chamfer distance. Traditionally, chamfer distance uses a distance transform to output a smooth cost function over an observation image and a fixed sized template. As a first step, we transform the rendering $I_R$ and data $I_D$ by computing the distance transform $D_f(I_D(i,j))$, which computes the value of closest point on the contour from every location in the image. We use the non-symmetric chamfer distance and use $I_R$ as the template since $I_D$ will typically have many outliers and will not be robust to variations. The likelihood $P(I_D|I_R)$ function can then be expressed as follows:
\[\text{Let }|I_R|_{+} = \sum_i^{row} \sum_j^{col} \delta_{I_R(i,j)} \hspace{5mm} \text{ and }  \hspace{5mm} D_f(I_D(i,j)) = \min_{m,n}||I_D(i,j)-I_D(m,n)|| \]
\[
\rho(I_D,I_R) = \frac{1}{|I_R|_{+}} \sum_i^{row} \sum_j^{col} DF(I_D(i,j))*I_R(i,j) \hspace{5mm}
P(I_D|I_R) = \frac{1}{\sqrt{2\pi\sigma_0^2}} \exp(-\frac{1}{2\sigma_0^2} \rho(I_D,I_R)^2)\]

We can now formulate the image interpretation task as approximately sampling the posterior distribution of scene elements $S_i$ given observations ($I_D$):
\begin{align}
P(S|I_D) \propto P(S) \delta_{g(\Theta*S)} P(I_D|I_R)
\end{align}

\subsection{Generative Model}
We demonstrate our framework on two challenging real world 3D parsing problems -- parsing generic 3D artifact objects and inferring fine 3D pose of humans from single images. The flexibility of our rich priors and the expressivity of the generic graphics simulator allows us to handle the extreme shape variability between the two problem domains with relative ease. For all generative models, we denote the affine transformations over an entire object as a latent matrix denoted by $S_L$. This matrix is a $4\times4$ affine matrix drawn from the the uniform distribution, where the range over translation variables is $Uniform(-1,1)$, range over scale is $Uniform(0.5,1.5)$ and range over rotation is $Uniform(-30,30)$ over all three axes independently.

\subsubsection{3D Object Parsing CAD Program}
Lathing and casting is a useful representation to express CAD models and is the main inspiration for our approach to modeling generic 3D artifacts. Given object boundaries in $\mathcal{B} \in \mathcal{R}^2$ space, we use a popular algorithm in graphics to lathe an object by taking a cross section of points, defining a medial axis for the cross section and sweeping the cross section across the medial axis by continuously perturbing with respect to $\mathcal{B}$. The space over $\mathcal{B}$ is extremely large due to variability in 3D shapes. We introduce a generative model consisting of Gaussian Processes (GPs) as a flexible non-parametric prior over object profiles for lathing (hereby referring to $\mathcal{B}$ by $f_{\mathcal{GP}}(x)$). The intermediate output from the graphics simulator is a mesh $G$ approximating the whole or part of a 3D object, which can be rendered to an image $I_R$ by camera re-projection.

The height $H$ of the object is sampled from a uniform distribution. 3D Objects can consist of several sub-parts. Without the loss of generality and for simplicity, we study objects with up-to two sub-parts and circular cross-section. We sample a cut $C$ along the medial axis of the object using the beta distribution, resulting in two independent GPs spanning the cut proportions. Since the smoothness and profile of 3D objects is a priori unknown, we need to do hyper-parameter inference on the bandwidths $L_1$ and $L_2$ of the covariance kernel of the GPs. The resulting points $f_{\mathcal{GP}}(x)$ from GPs are passed to the graphics simulator for lathing based mesh generation, which results in the generation of $I_R$. During inference, reconstructing 3D objects amounts to calculating the posterior $P(S=\{H,C,L_1,L_2,S_L\}|I_D)$. We show typical samples from the prior and an example inference trajectory in Figure ~\ref{fig:gp_quantitative}. The generative model can be formalized as follows:

\noindent\begin{minipage}{.60\textwidth}
\begin{align}
H &\sim Uniform(a_0,b_0)\\
C &\sim Rescaled\_beta(1,H) \\
L_1 &\sim a_1 + b_1 Beta(2,5) \text{ and } x_1 = [a_0, C] \\
L_2 &\sim a_1 + b_1 Beta(2,5) \text{ and } x_2 = [C+1, b_0]\\
k(x_p^i,x_p^j) &= \exp\Big(-\frac{(x_p^i-x_p^j)}{2 L_1^2} \Big), p\in\{1,2\}\\
f_{\mathcal{GP}}(x) &= \{\mathcal{GP}(0, k(x_p,x'_p))\}_{p=1}^2\\
I_R &= g_{LATHE}(S_L, f_{\mathcal{GP}}(x))
\end{align}
\end{minipage}
\begin{minipage}{.40\textwidth}
  \centering
  \includegraphics[width=22mm]{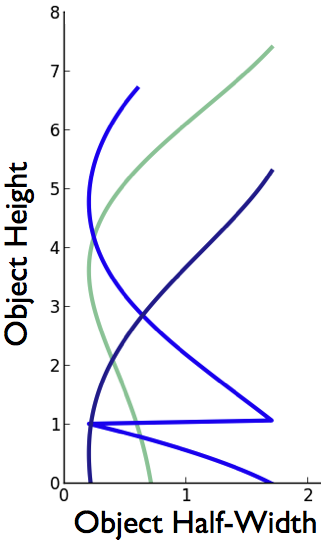}
  \captionof{figure}{Random draws from $f_{\mathcal{GP}}(x)$}
  \label{fig:figure}
\end{minipage}

\begin{figure*}
\centering
\includegraphics[width=45mm]{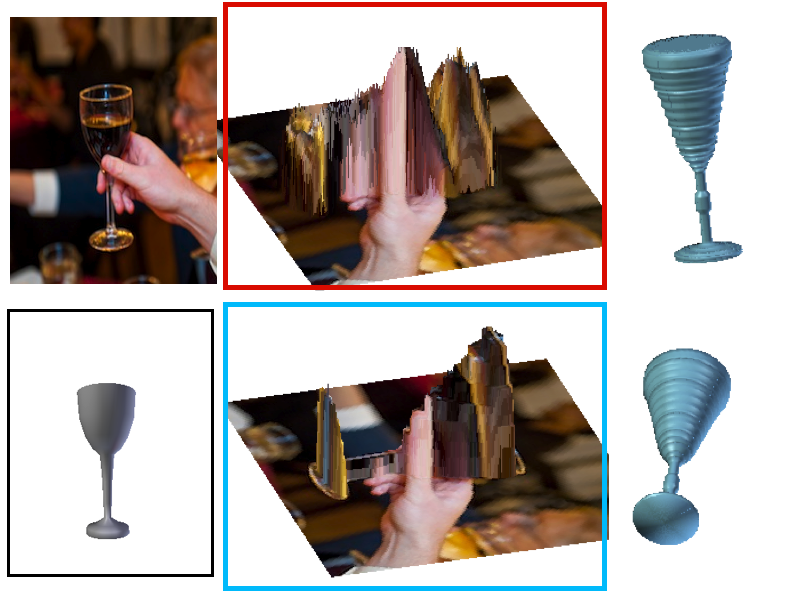}
\includegraphics[width=45mm]{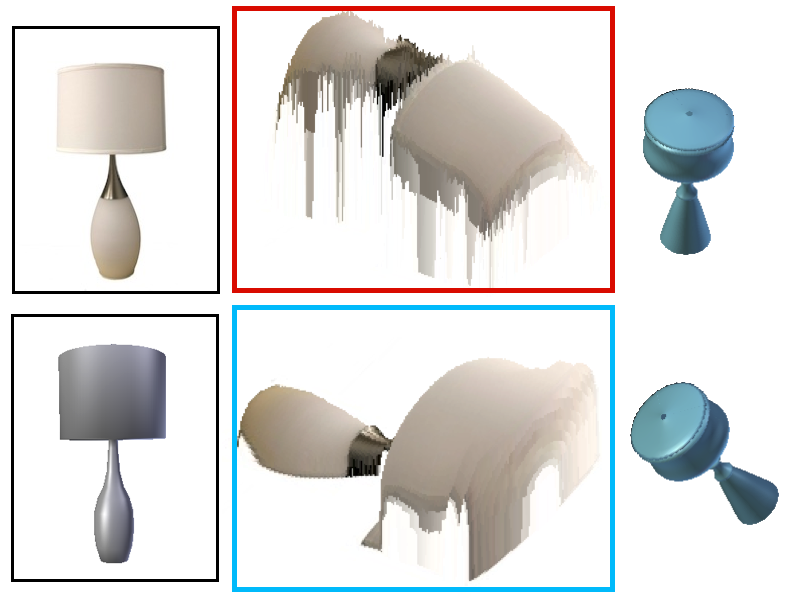}
\includegraphics[width=45mm]{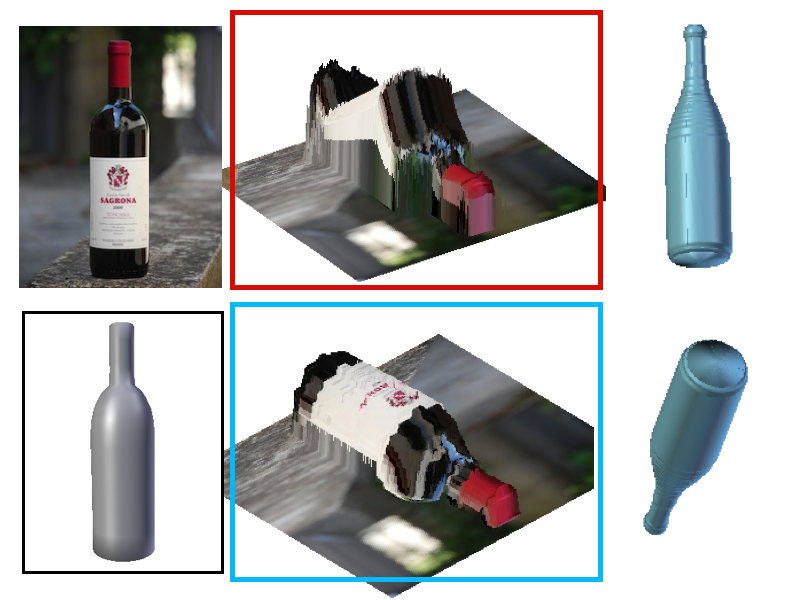}\\
\includegraphics[width=45mm]{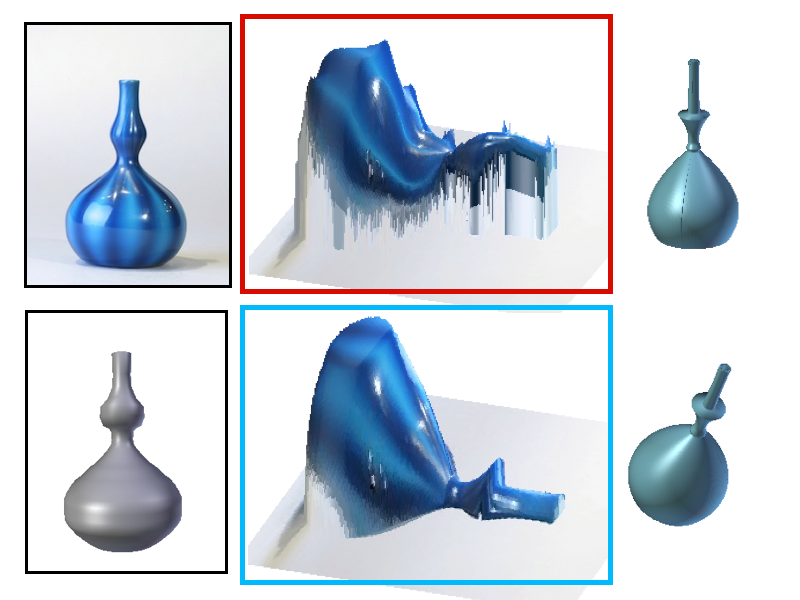}
\includegraphics[width=45mm]{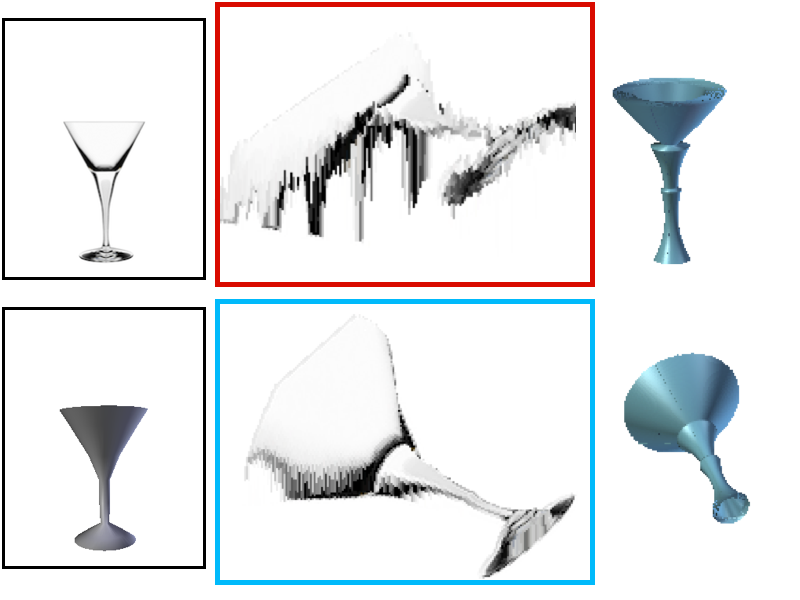}
\includegraphics[width=45mm]{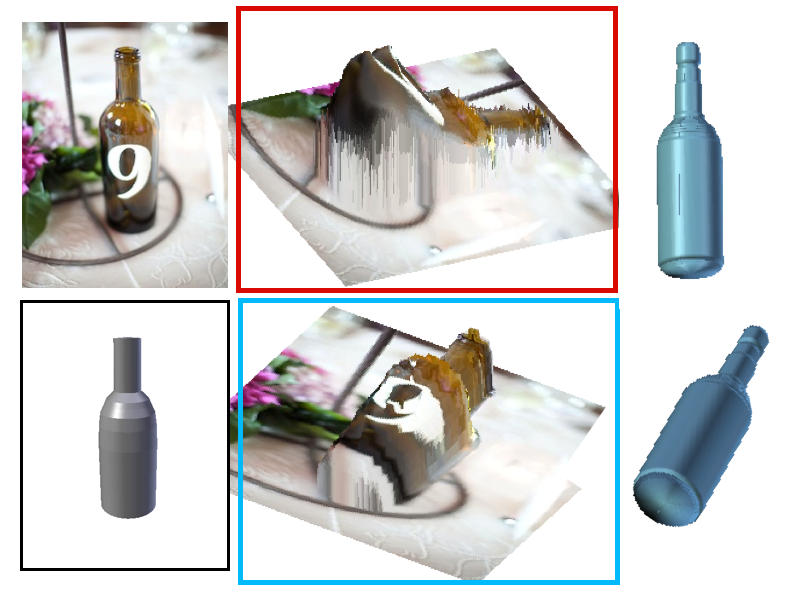}
\caption{{\bf Qualitative Results for 3D Object Parsing} in comparison to SIRFS~\cite{barron2013shape}. For every color image, the ground truth 3D model is shown in the box with black border below the color image. We super impose the depth buffer obtained from all methods on the original color images. The red box consist of results obtained from SIRFS and blue box depicts results obtained from our model -- both from roughly the same viewpoint. The blue 3D models rendered in different viewpoints are results obtained after collecting posterior samples from our model given the corresponding color images. As illustrated, our method gives dramatically better qualitative results on this challenging task -- suggesting that a strong 3D object prior can be beneficial for the problem of 3D visual reconstruction.}
\label{fig:gp_qualitative}
\end{figure*}

\subsubsection{3D Human Pose CAD Program}
We can naturally define a compositional model over the parts parameterized by the GP, such as an infinite mixture of GP experts or a hierarchical GP mixture model to learn 3D shapes such as human bodies. However, this is outside the scope of the current paper and is left for future work. In order to demonstrate deformable CAD programs, we designed a compositional mesh of the human body, where parts or vertex groups of the mesh approximate bones and joints. We use popular algorithm in graphics to generate armature for the resulting mesh and define priors over the armature control points.

The armature is a tree with the root node centered around the center of the mesh. The underlying armature tree is used as the medial axis to deform the mesh in a local part-wise fashion. Each joint/bone on the armature has a $4 \times 4$ affine matrix $S_L^i$ with scale $SC^i \sim \{\mathcal{U}_{t}(\mu_0,0.1)\}_{t \in x,y,z}$, rotation $R^i \sim \{\mathcal{N}_{t}(\mu_r,0.1)\}_{t \in x,y,z}$ and location $T^i \sim \{\mathcal{U}_{t}(a_t,b_t)\}_{t \in x,y,z}$ latent variables.The armature marked on the 3D mesh are depicted in Figure~\ref{fig:overview}b. Whenever the random choices in $S_L^i$ are re-flipped during inference, the change is propagated all the way to the root node or a pre-defined stopping node $S_L^j$ to smoothly evolve the 3D mesh. In order to assess coverage of our model, we show samples drawn from the prior in Figure~\ref{fig:overview}b and an illustrative run of the inference.

\section{Inference via Markov chain Monte Carlo}
Inference for inverting graphics simulators is intractable and therefore we resort to using Markov chain monte-carlo for approximate inference. Inference in our model is especially hard due to the following reasons: highly coupled variables, mix of discrete and continuous variables, many local minimas due to clutter, occlusion and noisy observations. For approximate inference to be possible in inverting the graphics simulator, we propose the following mixture of proposal kernels:

{\noindent {\bf Local Random Proposal:} Single site metropolis hastings moves on continuous variables and Gibbs moves on discrete variables. The proposal kernel is:}
\vspace{-2mm}
\begin{align}
q_{single}(S_{i}^{'}, S_i) &= \alpha_{single} P(S_{i}^{'})
\end{align}
{\noindent {\bf Block Proposal:} Affine latent variables such as the rotation matrix are almost always coupled with latent variables parameterizing the 3D mesh. Let us denote $\mathcal{A}(S) \in S$ to be a set of latents belonging to the same affine transformation and $\{S_l\}_{l=1}^L \in S$ to be all other non-affine latents. In order to allow the affine latent groups to mix, we use the following blocked proposal:}
\vspace{-2mm}
\begin{align}
q_{block}(\mathcal{A}(S)',\{S'_l\}_{l=1}^L,\mathcal{A}(S),\{S_l\}_{l=1}^L) &= \alpha_{block} \prod_{S' \in A(S)'}P(S') \prod_{l=1}^L P(S'_l)
\end{align}

{\noindent {\bf Discriminative Proposal:} Despite asymptotic convergence, inference often gets suck in local minimas due to occlusion, clutter and noisy observations. Often times, the only way to escape a local minima is if the sampler could make precise changes to large number of latent variables at the same time. We follow a strategy similar to ~\cite{jampani2014informed} to learn data-driven proposals:}
\vspace{-2mm}
\begin{align}
&\text{Sample from the prior or use annotated training data: } \{I_D^n, S^n\}_{n=1}^N\\
&\text{Given } I_D \text{, find K-nearest } \{S^k\}_{k=1}^K \text{ in some feature space } \phi(I_D^n):\rightarrow R^M\\
&q_{data} = \alpha_{data} \text{\it KDE}(\{S^k\}_{k=1}^K)
\vspace{-1mm}
\end{align}

{\noindent {\bf Local Gradient Proposal:} Since our likelihood function is smooth for continuous variables and often times there is coupling between these variables, we can exploit gradient information by constructing a Hamiltonian kernel $q^{HMC}(S_{c}^{'}, S_{c})$, where $S_{c}$ denotes all the continuous scene variables.}

During inference, we approximate $P(S|I_D)$ by accepting or rejecting with the following ratio:
\begin{align}
\alpha_{d}(S' \rightarrow S) &= \frac{P(I_D|g(\Theta*S') \prod_{i}(S'_i) q^{*}(S_i' \rightarrow S_i))}{P(I_D|g(\Theta*S) \prod_{i}P(S_i) q^{*}(S_i \rightarrow S_i'))} \\
\text{, where } q^{*} &=\alpha_{single} q_{single} + \alpha_{block} q_{block} + \alpha_{data} q_{data} + \alpha_{hmc} q_{hmc}
\end{align}

\begin{figure*}
\centering
\includegraphics[width=60mm]{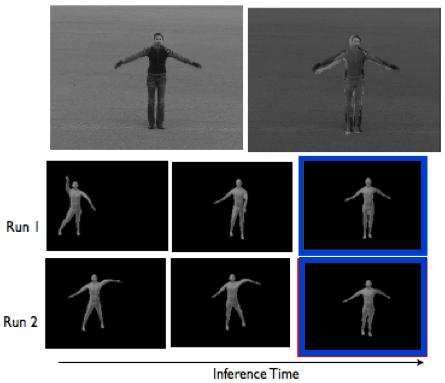}
\includegraphics[width=65mm]{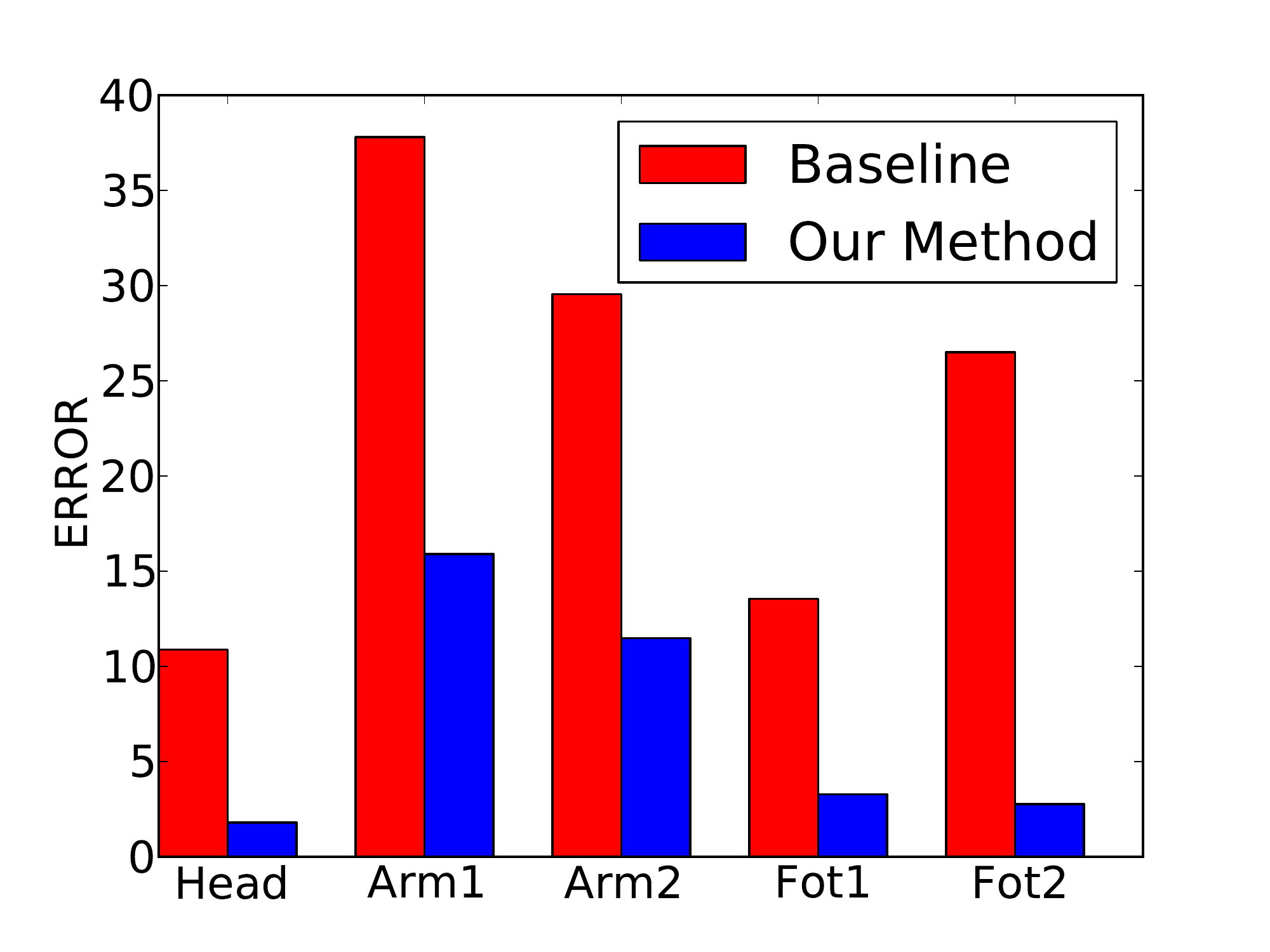}
\caption{{\bf Quantitative Results for 3D Human Pose Estimation:} (a) Illustration of inference on two independent runs for a given test image. Even without making any assumptions on affine transformations of the human body such as scale or location, inference converges reliably to the posterior. The resulting 3D mesh is projected and super-imposed onto the test image to show localization accuracy. (b)
We quantitatively evaluate the performance on a dataset collected from various sources such as KTH~\cite{schuldt2004recognizing}, LabelMe~\cite{russell2008labelme} images with significant occlusion in the ``person sitting'' category and the Internet. Our model significantly outperforms the current state of the art DPM human pose detector~\cite{yang2011articulated}}
\label{fig:pose_quantitative}
\end{figure*}

\section{Experiments}
\subsection{3D Object Parsing}
We collected a small dataset of about 20 3D objects from the internet and demonstrate superior results over the current state of the art model for single object reconstruction~\cite{barron2013shape}. As compared to the baseline, our model does not require pre-segmented objects as we do joint inference over affine transformations and mesh elements. Evaluating 3D reconstruction is challenge for single images. We asked CAD experts to manually generate CAD fits to images in blender and evaluated both the approaches by calculating Z-MAE score (shift-invariant surface mean-squared error) and N-MSE (mean-squared error over normals). As described in Figure~\ref{fig:gp_quantitative}a, our model has a much lower Z-MAE and N-MSE score than SIRFS without utilizing pre-segmented objects. Moreover, most of the error in our approach can be attributed to slight misalignment of the 3D parse from our model with respect to ground truth (SIRFS has an inherent advantage in this metric since objects are pre-segmented). Most importantly, as demonstrated in Figure~\ref{fig:gp_qualitative}, our approach has dramatically better qualitative 3D results as compared to SIRFS. Figure~\ref{fig:gp_quantitative}b shows typical trajectory of the inference from prior to the posterior on a challenging real world image. In future work, we hope to naturally extend our model to handle arbitrary cross sections (same GP based model) and a much more structured hypothesis space to represent Gaussian Process based object parts.

\subsection{3D Human Pose Estimation}
We collected a small dataset of humans performing a variety of poses from sources such as KTH~\cite{schuldt2004recognizing}, LabelMe~\cite{russell2008labelme} images with significant occlusion in the ``person sitting'' category and the Internet (about 50 images), and demonstrate superior results in comparison with the current state-of-the-art Deformable Parts Model (DPM) for pose estimation~\cite{yang2011articulated}. We project the 3D pose obtained from our model to 2D key-points (such as head, left arm etc.) for comparison with the baseline. As shown in Figure~\ref{fig:pose_quantitative}, we outperform the baseline by a significant margin.

\begin{figure*}
\centering
\includegraphics[height=90mm]{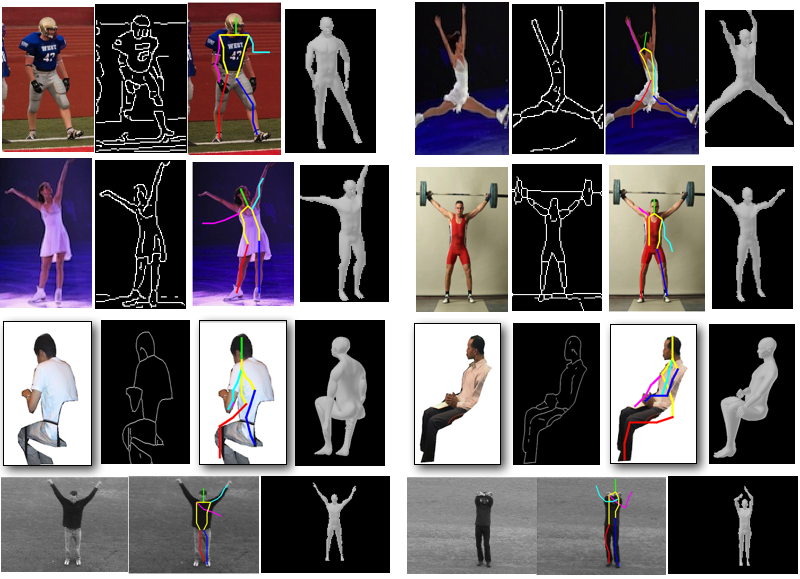}
\caption{{\bf Qualitative Results for 3D Pose Estimation} in comparison with the DPM Pose detector~\cite{yang2011articulated}. The first column for every set of results contain the test(color) image, followed by the contour image. DPM results are shown as stick figures on the original image. Results from our 3D model are rendered on a black background. Our method consistently outperforms the baseline. Many of our model's failure cases are due to errors in localizing hands, which is primarily due to noisy or low resolution contour map. In the future, better bottom-up features that give finer contours and other mid-level representations such as texture descriptors~\cite{portilla2000parametric} are interesting avenues for improving the results.}
\label{fig:pose_qualitative}
\end{figure*}

As shown in Figure~\ref{fig:pose_qualitative}, images with people sitting and heavy occlusion are very hard for the discriminative model to get right -- mainly due to ``missing'' observation signal -- making a strong case for a model based approach like ours which seems to give reasonable parses. Most of our model's failure cases as shown in Figure~\ref{fig:pose_qualitative} and ~\ref{fig:pose_quantitative}b, are in inferring arm position; this is typically due to noisy and low quality contour maps around the arm area due to its small size. In the subsequent section, we will utilize the strengths of the bottom-up DPM pipeline to learn a better strategy for doing probabilistic inference.

\subsubsection{Discriminative Data-driven Proposals}
\begin{figure}[t]
\centering
\includegraphics[width=120mm]{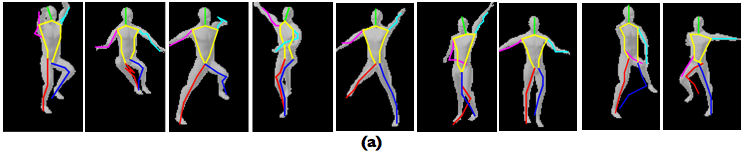}\\
\includegraphics[width=45mm]{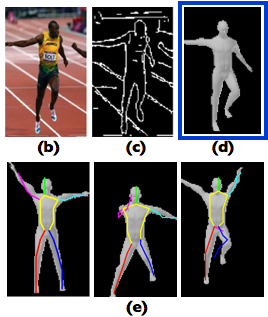}
\includegraphics[width=65mm]{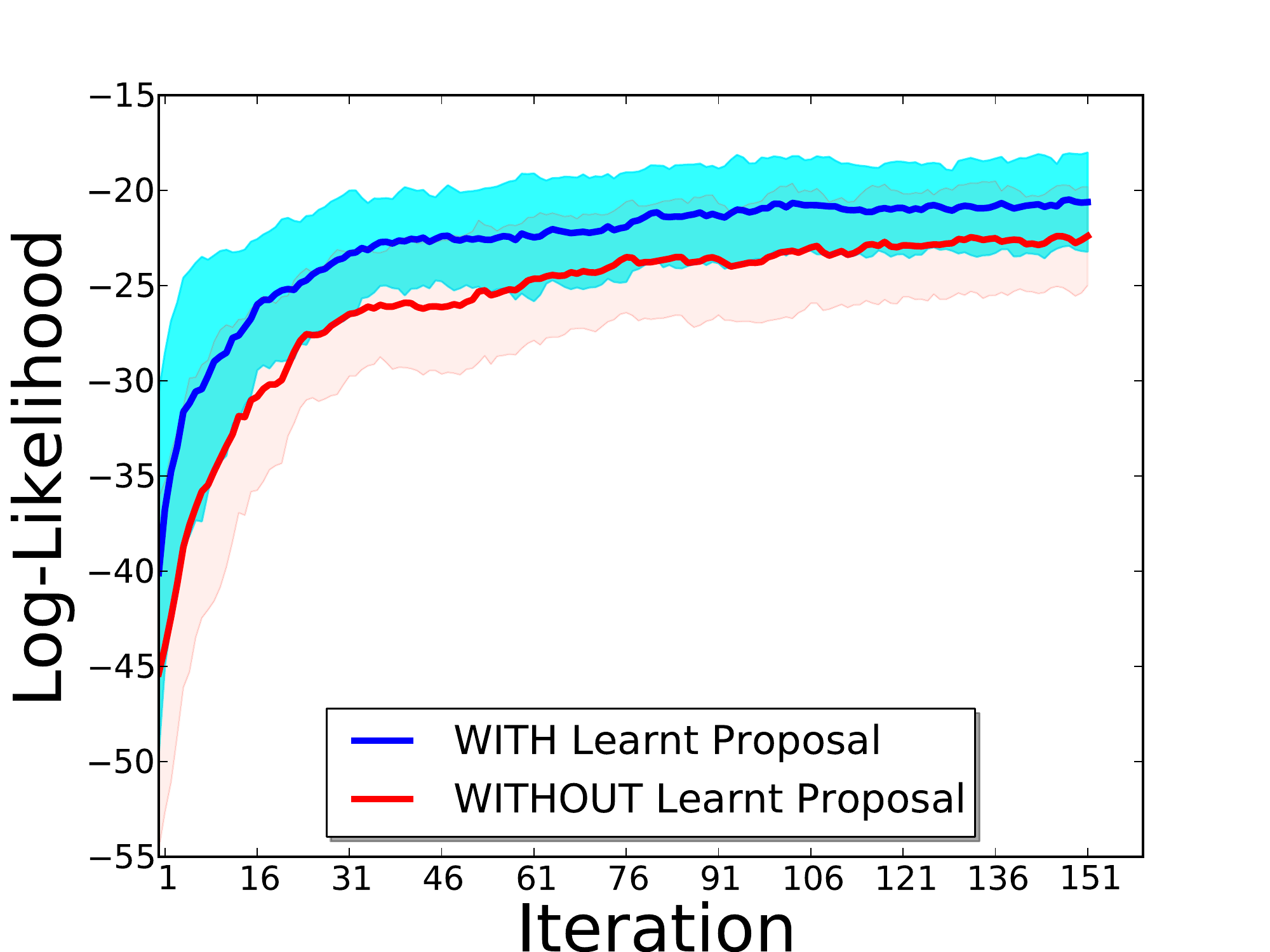}\\
\caption{{\bf Discriminative Proposal Learning:} {\bf(a)} Following procedure highlighted in Section 3, We generate large number of samples from the prior and fit the DPM pose detector~\cite{yang2011articulated} to the resulting rendered images (stick figure depict DPM result).We treat the bounding box outputs from DPM as out features. {\bf(b)} Given a color image, we calculate the features using DPM pose detector and retrieve K-nearest neighbors from data generated in (a). Given the neighbors, we fit a kernel density estimator (KDE) to its latent variables to get the learned proposal. {\bf(d)} A sample parsed result with proposal learning. {\bf(e)} Samples drawn from the KDE given the color image (b) are semantically close to the posterior. Inference fine-tunes the solution to a better fit via other proposal kernels in our algorithm. As shown on the log-l plot, we run about 100 independent chains with and without the learned proposal. Inference with learned proposal consistently outperforms baseline in terms of both speed and accuracy.}
\label{fig:pose_proposal}
\end{figure}

To aid inference in 3D human pose estimation, we also explore the use of discriminatively trained predictors that act as sample generators in our inversion pipeline. We sample a large number of samples from the prior and fit the pose estimated on the rendered images using the feed-forward DPM pose detection pipeline. During inference, we fit the DPM pose model on the real test image, find K-nearest neighbors from the sampled dataset in the space of DPM pose parameters and a local density estimator (KDE) to generate a discriminatively trained proposal. Intuitively, the feed-forward pathways rapidly initializes the latent variables to a reasonable state, leaving further fine tuning up to inference pipeline. This effect is confirmed by Figure~\ref{fig:pose_proposal}, where the about 100 independent Markov chains are ran with and without discriminative proposals. The results consistently favors the runs with discriminative learning both in terms of accuracy and speed.

\section{Discussion}

We have shown that it is possible to solve challenging, real-world 3D vision problems by approximately Bayesian inversion of probabilistic CAD programs. Shape variability is addressed using 3D modeling tools from computer graphics and with nonparametric Bayesian statistics to deal with unknown shape categories. Appearance variability is handled by comparing renderings to real-world images in a feature space based on mid-level representations and distance metrics from computer vision. Inference is handled via a mixtures of Hamiltonian Monte Carlo, standard single-site and locally blocked Metropolis-Hastings moves. This approach yields quantitative and qualitative performance improvements on 3D human pose estimation and object reconstruction as compared to state-of-the-art baselines, with only moderate computational cost. Additionally, data-driven proposals learned from synthetic data, incorporating representations from human pose detectors, can be used to improve inference speed.

Several research directions seem appealing on the basis of these results. Scaling in object and scene complexity could be handled by incorporating object priors based on hierarchical shape decompositions or infinite mixtures of piecewise GPs. The explicit 3D scenes could be augmented with physical information, with kinematic and physical constraints integrated via undirected potentials; it may begin to be practical to build systems that perform rich physical reasoning directly from real-world visual data. Image comparison could be improved by using richer appearance models such as Epitomes~\cite{papandreou2014modeling}. It could also be fruitful to experiment directly modeling reflectance and illumination via an approach like SIRFS~\cite{barron2013shape}, though choosing the right resolution for comparison may be difficult. It seems natural to explore richer bottom-up proposal mechanisms that integrate state-of-the-art discriminative techniques, including modern artificial neural networks~\cite{KrizhevskySH12}. Many of these directions, as well as the exploration of alternative inference strategies, may be simplified by implementation as generative probabilistic graphics programs atop general-purpose probabilistic programming systems~\cite{mansinghka2014venture}.

The generative, approximately Bayesian approach to vision has a long way to go before it can begin to compete with the flexibility and maturity of current bottom-up vision pipelines, let alone their computational efficiency. Basic design trade-offs in modeling and inference are not yet well understood, and to make a 3D scene parser that performs acceptably on the object recognition challenges like PASCAL may be a good proxy for the general problem of visual perception. Despite these limitations, however, it in some ways offers a clearer scaling path to rich percepts in uncontrolled settings. Our results suggest it is now possible to realize some of this potential in practice, and not only produce rich, representationally explicit percepts but also obtain good quantitative performance. We hope to see many more illustrations of this kind of approach in the future.

\section{Acknowledgments}
We thank Peter Battaglia, Daniel Selsam, Owain Evans, Alexey Radul and Sam Gershman for their valuable feedback and discussions. Tejas Kulkarni was graciously supported by the Henry E. Singleton Fellowship. Partly funded by the DARPA PPAML program, grants from the ONR and ARO, and Google's "Rethinking AI" project.
\bibliographystyle{abbrv}
\small{\bibliography{pcad_nips}}

\end{document}